\documentclass[conference]{IEEEtran}
\IEEEoverridecommandlockouts
\usepackage{cite}
\usepackage{amsmath,amssymb,amsfonts}
\usepackage{algorithmic}
\usepackage{graphicx}
\usepackage{textcomp}
\usepackage{xcolor}

\usepackage{booktabs}
\usepackage{multirow}
\usepackage{amsmath}
\usepackage{tikz}
\newcommand*\circled[1]{\raisebox{0.325ex}{\tikz[baseline=(char.base)]{
            \node[shape=circle,draw,inner sep=0.2pt, scale=0.75] (char) {\textbf{\fontsize{8}{9}\selectfont #1}};}}}

\usepackage{threeparttable}
\usepackage{pifont}
\usepackage{url}
\usepackage{enumitem}
\usepackage{makecell}
\usepackage{hyperref}

\def\BibTeX{{\rm B\kern-.05em{\sc i\kern-.025em b}\kern-.08em
    T\kern-.1667em\lower.7ex\hbox{E}\kern-.125emX}}
\begin{document}

\title{Detect, Investigate, Judge and Determine:\\A Knowledge-guided Framework for Few-shot Fake News Detection
\thanks{{\bf *}Corresponding author.}
}

\author{
{\bf Ye Liu$^{1,2}$}, {\bf Jiajun Zhu$^{1,2}$}, {\bf Xukai Liu$^{1,2}$}, {\bf Haoyu Tang$^{1,2}$} \\{\bf Yanghai Zhang$^{1,2}$}, {\bf Kai Zhang$^{1,2}$},  {\bf Xiaofang Zhou$^{3}$}, {\bf Enhong Chen$^{1,2,}$*} \\
        $^1$ University of Science and Technology of China,
        $^2$ State Key Laboratory of Cognitive Intelligence\\
        $^3$ The Hong Kong University of Science and Technology\\
        {yeliu.liuyeah@gmail.com}, \{jiajunzhu, chthollylxk, haoyu\_t, yhzhang0612\}@mail.ustc.edu.cn\\
        \{kkzhang08, cheneh\}@ustc.edu.cn, zxf@ust.hk\\
}

\maketitle

\begin{abstract}
Few-Shot Fake News Detection (FS-FND) aims to distinguish inaccurate news from real ones in extremely low-resource scenarios. This task has garnered increased attention due to the widespread dissemination and harmful impact of fake news on social media. Large Language Models (LLMs) have demonstrated competitive performance with the help of their rich prior knowledge and excellent in-context learning abilities. However, existing methods face significant limitations, such as the Understanding Ambiguity and Information Scarcity, which significantly undermine the potential of LLMs. To address these shortcomings, we propose a Dual-perspective Knowledge-guided Fake News Detection (DKFND) model, designed to enhance LLMs from both inside and outside perspectives. Specifically, DKFND first identifies the knowledge concepts of each news article through a Detection Module. Subsequently, DKFND creatively designs an Investigation Module to retrieve inside and outside valuable information concerning to the current news, followed by another Judge Module to evaluate the relevance and confidence of them. Finally, a Determination Module further derives two respective predictions and obtain the final result. Extensive experiments on two public datasets show the efficacy of our proposed method, particularly in low-resource settings.
\end{abstract}


\section{Introduction}
\label{sec:introduction}
Fake News Detection (FND), aiming to distinguish between inaccurate news and legitimate news, has garnered increasing importance and attention due to the pervasive dissemination and detrimental effects of fake news on social media platforms\cite{shu2017fake}. Few-Shot Fake News Detection (FS-FND), as a subtask of fake news detection, endeavors to identify the fake news by leveraging only $K$ instances per category ($K$-shot) in the training phase\cite{hu2024bad,gao2021making,ma2023kapalm}. 


Generally, fake news detection can be framed as a binary classification problem and addressed using various classification models. In the early stage, researchers primarily employ machine learning or deep learning algorithms to represent and classify candidate news articles\cite{horne2017just,jiang2022fake}. More recently, with the rise of Large Language Models (LLMs), FSFND has been effectively addressed through the in-context learning technology, which is particularly prevalent in few-shot settings\cite{hu2024bad,boissonneault2024fake}. Among them, Hu and Wang et al.\cite{hu2024bad,wang2024explainable,teo2024integrating} were pioneers in investigating the potential of LLMs in this field. 

\begin{figure}[t]
    \centering
    \includegraphics[width=8cm]{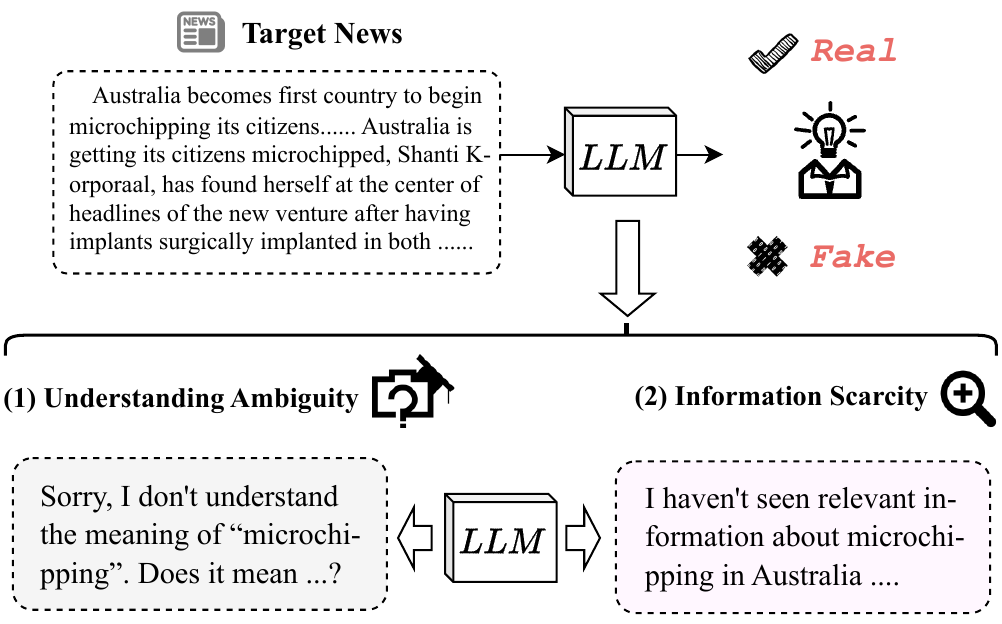}
    \caption{An example of fake news detection and the limitations of existing LLM-based methods.}
    \vspace{-5mm}
    \label{fig:intro}
\end{figure}
However, despite these advancements, existing LLM-based methods face inherent challenges. They often rely on directly prompting LLMs to judge the authenticity of the given news, which often exceed the capabilities of these models, particularly those relatively small LLMs (e.g., 7B parameters) that are commonly used in practical applications. An example of this can be observed in Figure~\ref{fig:intro}, which presents a news about microchip developments. Existing LLM-based approaches encounter two principal challenges: \textbf{(1) Understanding Ambiguity}: LLMs may fail to understand and grasp the core meaning conveyed in the news, especially when it contains subtle nuances or domain-specific jargon, thereby straining the detection process. \textbf{(2) Information Scarcity}: Given the dynamic and fast-paced nature of news, the training corpus of LLMs is often outdated, leading to a lack of relevant and up-to-date information during the detection process.

To this end, this paper proposes a novel approach to address the two aforementioned issues. Specifically, to mitigate the Understanding Ambiguity problem, we aim to extract valuable insights from an inside perspective by retrieving similar samples from the training set, thereby enhancing the comprehension of key concepts in the target news. Simultaneously, to tackle the Information Scarcity problem, we employ an external search engine to gather relevant information about the news from the web. This design integrates more real-time data, effectively overcoming the limitation of information obsolescence. Importantly, in the whole process, an external knowledge graph is employed to enhance the reliability and information relevance.

More specifically, we design a Dual-perspective Knowledge-guided Fake News Detection (DKFND) model. DKFND comprises four key components:
(a) A Detection Module: This module leverages a knowledge graph to identify relevant knowledge concepts within the text, forming the basis for queries used in the subsequent modules.
(b) An Investigation Module: It investigates more valuable information related to the target news, which comes from both inside (i.e., training set) and outside (i.e., search engine) perspectives.
(c) A Judge Module: With the help of knowledge graphs and LLMs, this module assesses the relevance and authenticity of investigated informations.
(d) A Determination Module: It designs specialized prompts to guide LLMs in generating predictions and explanations from both two perspectives. Subsequently, it makes a final decision with high confidence, especially in cases where the inside and outside perspectives conflict.

We conduct extensive experiments on two publicly available datasets, where the experimental results demonstrate the superiority of our proposed method. Our codes is avaiable via \url{https://github.com/liuyeah/DKFND}.


\section{Related Work}
\subsection{Few-Shot Fake News Detection}

Generally, fake news detection can be defined as a binary classification problem and addressed by a variety of classification models. Initially, researchers mainly rely on feature engineering and machine learning algorithms\cite{horne2017just}. As computing power and data availability have increased, significant improvements have been made with the help of various deep learning algorithms and Pre-trained Language Models (PLMs). For instance, Ghanem et al.\cite{ghanem2021fakeflow} combined lexical features and a Bi-GRU network to achieve accurate fake news detection. Jiang et al.\cite{jiang2022fake} introduced the Knowledgeable Prompt Learning (KPL), a novel framework that integrates prompt learning with fake news detection, achieving state-of-the-art performance in few-shot settings.

Additionally, researchers have also recognized the importance of external knowledge to complement traditional fake news detection methods. News articles often contain references to entities, events, and facts that are external to the article itself, which makes external knowledge integration crucial for accurate detection. For instance, Dun and Ma et al.\cite{dun2021kan,ma2023kapalm,hu2021compare} utilized knowledge graphs to enrich entity information and structured relation knowledge, leading to more precise news representations and improved detection performance. Meanwhile, Huang et al.\cite{huang2023faking} adopted a data augmentation perspective, proposing a novel framework for generating more valuable training examples, which has proven to be beneficial in detecting human-written fake news.

More recently, with the advent of large language models, many researchers are exploring few-shot fake news detection through in-context learning and data augmentation technologies\cite{hu2024bad,wang2024explainable,teo2024integrating}. For example, Hu et al.\cite{hu2024bad} explored the role of LLMs in fake news detection, proposing an Adaptive Rationale Guidance (ARG) network that combines traditional detection techniques with the generative capabilities of LLMs. However, many of these approaches rely on directly prompting the LLM to classify the news as fake or real, without fully utilizing the broader potential of LLMs to understand and reason about the news context. More importantly, most of them are significantly limited by the aforementioned two shortcomings, particularly in the Information Scarcity problem.


\subsection{Large Language Models \& RAG}

Large Language Models (LLMs) such as GPT-4, LLama-3, and others have revolutionized natural language processing, achieving impressive results across a variety of tasks, including text classification, summarization, information extraction, and fake news detection\cite{hoffmann2022training,openai2023gpt4,llama3modelcard,tunstall2023zephyr}. One of the key innovations introduced by LLMs is in-context learning, a paradigm in which the model learns from a few examples presented in the context of the task, allowing it to perform few-shot learning without the need for explicit retraining\cite{brown2020language,hu2024bad,wang2024explainable}. 

In parallel, the integration of Retrieval-Augmented Generation (RAG) has emerged as a promising approach to enhance the capabilities of LLMs. RAG models, such as those proposed by\cite{lewis2020retrieval}, combine the generative power of LLMs with the retrieval of relevant documents or knowledge from an external corpus. This approach not only improves the factual accuracy of generated content but also allows for more informed and contextually appropriate responses in tasks like question answering and knowledge-intensive applications\cite{izacard2021leveraging,borgeaud2022improving,izacard2023atlas}. 

In the context of few-shot fake news detection, incorporating external knowledge retrieval has shown promising results. Jiang et al.\cite{jiang2022fake} were among the first to integrate retrieval-based methods with LLMs for fake news detection. More specifically, by incorporating knowledge from Wikidata into the prompt representation process, they devised a knowledgeable prompt learning framework that significantly improved the detection performance. Similarly, other studies have optimized external retrieval mechanisms to enhance the performance of LLMs, allowing more context-aware and accurate prediction results\cite{borgeaud2022improving,izacard2023atlas}.

\section{Problem Statement}
Generally, fake news detection can be framed as a binary classiﬁcation problem, wherein each news article is classified as either real ($y = 0$) or fake ($y = 1$)\cite{dun2021kan}. Formally, each piece of news $S$ is composed of a sequence of words, i.e., $S = \{s_1, s_2, ..., s_n\}$, encompassing its title, content text, and relevant tweets. The goal is to learn a detection function $F: F(S) \Longrightarrow y$, where $y \in \{0,1\}$ denotes the ground-truth label of news.

In the few-shot settings, adhering the strategy employed in\cite{gao2021making,ma2023kapalm}, we randomly sample $K$ instances per category ($K$-shot)\footnote{This implies that for a $K$-shot fake news detection setting, the number of training instances is $2K$.} for the training phase. The entire test set is preserved to ensure the comprehensiveness and effectiveness of evaluation.

\section{The DKFND Model}
\label{section:model}
\begin{figure*}[t]
    \centering
    \includegraphics[width=\textwidth]{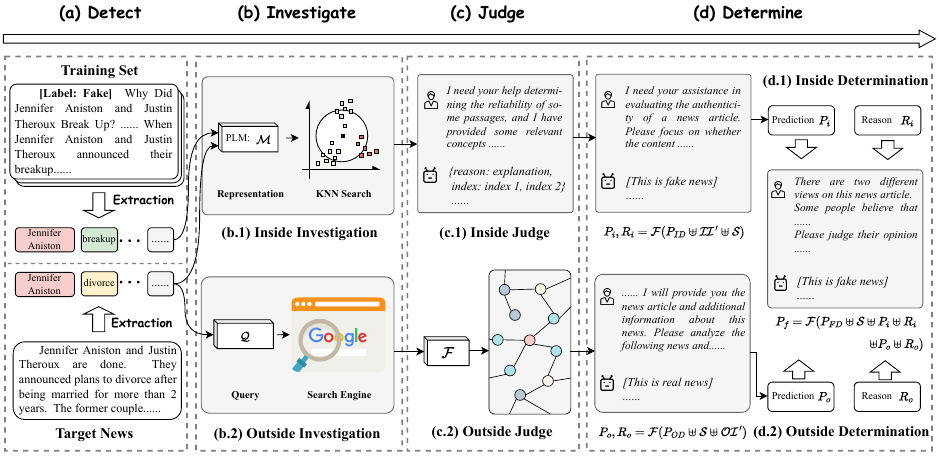}
    \caption{The architecture of our DKFND model. It includes four sequentially connected parts: (a) Detection Module; (b) Investigation Module; (c) Judge Module; (d) Determination Module.}
    \vspace{-5mm}
    \label{fig:architecture}
\end{figure*}

In this section, we will introduce the technical details of DKFND model, as illustrated in Figure~\ref{fig:architecture}. 

\subsection{Detection Module}
\label{section:detection}
In this module, we aim to identify the key information contained in the given news article, which will serve as the query for the subsequent modules. Specifically, we employ the \textit{Spacy} algorithm\cite{honnibal2017spacy} to extract relevant knowledge concepts $C = \{c_1, c_2, ..., c_m\}$.

Subsequently, we construct prompts to guide the LLM to select the top-$N
$ most critical knowledge concepts, which are expected to answer the question: \textit{“How relevant these knowledge concepts are to the given news?”}:
\begin{equation}
    C' = \mathcal{F} (P_{detection}, S, C),
\end{equation}
where $\mathcal{F}$ represents the LLM, and $P_{detection}$ denotes the prompt for the in-context learning. More detailed description can be found in Appendix~\ref{appendix:promot}.

\subsection{Investigation Module}
\label{section:investigation}
In this module, we aim to investigate the relevant information, from two perspectives: Inside Investigation and Outside Investigation.

\subsubsection{Inside Investigation.}
\label{section:inside_investigation}
To address the Understanding Ambiguity problem introduced in Section~\ref{sec:introduction}, we retrieve effective demonstrations to enhance the LLM's understanding during the in-context learning process\cite{liu2022makes,rubin2022learning}. 

Specifically, we first concatenate the extracted $N$ concepts of each news, and then utilize the pre-trained language model $\mathcal{M}$ to obtain the representation of these concepts $\{c_1, ..., c_N\}$:
\begin{equation}
\begin{aligned}
    Q = c_1 &\uplus c_2 \uplus ... \uplus c_N, \\
    H &= \mathcal{M} (Q),
\end{aligned}
\end{equation}
where $\uplus$ represents the concatenation operation. The derived representation $H$ is used to represent each news sample. Along this line, we can further obtain the representation and label pairs $(H_i, l_i)$ for the training set, which constitute a datastore, denoted as $D$. 

Subsequently, when inferring a candidate news $j$, we employ the $k$-Nearest Neighbors ($k$NN) search method\cite{khandelwal2019generalization} to retrieve valuable samples from the training set. In detail, we use the representation $H_j$ of news $j$ to query the datastore $D$ according to the euclidean distance. Then, based on the computed distance, we select the nearest $k$ positive and negative news samples, respectively: 
\begin{equation}
    \mathcal{II} = \{ U_{positive}, U_{negative} \}.
\end{equation}
As a consequence, we obtain the inside investigation outcome $\mathcal{II}$, comprising $2k$ instances.

\subsubsection{Outside Investigation.}
\label{section:outside_investigation}
In response to the Information Scarcity problem, we further retrieve additional real-time information from external sources. Inspired by\cite{yoran2023answering,paranjape2023art}, we implement a retriever based on the google search engine, using the SerpAPI service\footnote{\url{https://serpapi.com/}. You can refer to Appendix~\ref{appendix:outside_search} for a searching example.}. 

Specifically, based on the extracted concepts $\{c_1, c_2, ..., c_N\}$ in Section~\ref{section:detection}, we first concatenate them to construct the initial query $\mathcal{Q} = c_1 \uplus c_2 \uplus ... \uplus c_N$. Then, following the strategy proposed in\cite{yoran2023answering}, we further format the search queries as “\texttt{en.wikipedia.org} $\mathcal{Q}$”, with the Wikipedia domain preceding the intermediate question. We return the top-\textit{L} evidence retrieved by Google. And all retrieved evidence sentences are prepended to the outside investigation outcome, denoted as $\mathcal{OI} = \{G_k, k = 1,2,..,L \}$.

\subsection{Judge Module}
\label{section:judge}
We design this module to assess the relevance and authenticity of the retrieved information.

\subsubsection{Inside Judge}
\label{section:inside_judge}
In the Inside Investigation module (Section~\ref{section:inside_investigation}), we employ the \textit{k}NN search method to retrieve relevant documents $\mathcal{II}$ from the training set. However, the retrieved content may still contain irrelevant information, potentially introducing noise and interfering with the prediction process.

To address this issue, we propose an inside judge method that leverages the knowledge concepts identified in the Detection Module. Specifically, we design prompts to guide the LLM in further refining the retrieved documents by selecting the top- $m$ most relevant samples. Specifically:
\begin{equation}
\begin{aligned}
    U_{positive}' &= \mathcal{F} \ (P_{IJ}\ \uplus\ U_{positive}\ \uplus \ C'),\\
    U_{negative}' &= \mathcal{F} \ (P_{IJ}\ \uplus\ U_{negative}\ \uplus \ C'),
\end{aligned}
\end{equation}
where $P_{IJ}$ denotes the prompt and more detailed can be found in Appendix~\ref{appendix:promot}. Finally, the selected inside investigated information can be denoted as $\mathcal{II}' = \{ U_{positive}', U_{negative}' \}$.

\subsubsection{Outside Judge}
\label{section:outside_judge}
For the Ouside Investigation module (Section~\ref{section:outside_investigation}), although the retrieved information is obtained via the Google API, ensuring the relevance, a major limitation lies in the uncertainty of its authenticity and accuracy. To address this, we incorporate a knowledge graph to verify the authenticity of the retrieved outside information.

\paragraph{Concept Triple Identification.}
In this part, we aim to identify the knowledge concept triples from each document $G_k$ in $\mathcal{OI}$. Specifically, we first employ \textit{Spacy} algorithm to recognize the knowledge concept set $\mathcal{E}_k = \{e_i\ |\ i = 1,2...\}$ from $G_k$. Then, we design prompts to guide the LLM to compose triples:
\begin{equation}
    \mathcal{T}_k = \mathcal{F} \ (P_{OJ}\ \uplus\ G_k\ \uplus \ \mathcal{E}_k),
\end{equation}
where $P_{OJ}$ denotes the prompt, and $\mathcal{T}_k = \{T_k^l|T_k^l = (h_k^l, r_k^l, t_k^l), l=1,2... \}$. Subsequently, to refine the relation $r_k^l$, we utilize the semantic matching method to find a closest relation in KGs. For convenience, we still use $\mathcal{T}_k$ to denote the refined concept triples.

\paragraph{Triple Authenticity Assessment.}
In this part, we further employ the KG embedding model: ComplEx\cite{trouillon2016complex}, which is pretrained on Wikidata through LibKGE\cite{broscheit2020libkge}, to assess the authenticity of each triple $T_k^l$ in $\mathcal{T}_k$:
\begin{equation}
    Sc(T_k^l) = f(h_k^l, r_k^l, t_k^l),
\end{equation}
where $f(h, r, t)$ denotes the score function of the pretrained ComplEx model. $Sc(T_k^l)$ measures the possibility of $T_k^l$ appearing in given KG. Further, we compute the average score of all triples occuring in the document $G_k$:
\begin{equation}
    Score(G_k) = Avg(Sc(T_k^l), \ T_k^l \in \mathcal{T}_k).
\end{equation}

\paragraph{Document Selection.}
Relying on the authenticity score of $G_k \in \mathcal{OI}$, we could finally obtain the selected documents. Specifically, considering the ranking design of Google search\cite{patil2021comparative}, we first choose the top-1 document $G_1$ in $\mathcal{OI}$. Then we select top n-1 documents from the remaining L-1 documents, through the authenticity score $Score(G_k)$. Finally, we obtain the selected outside investigated information $\mathcal{OI'} = \{G_k, k = 1,2,..,n \}$.

\subsection{Determination Module}
\subsubsection{Inside Determination}
\label{sec:inside_determination}

After obtaining the effective demonstrations $\mathcal{II'}$ from Inside Judge, we design prompts to provide the essential information to the LLM, thereby generating the inside prediction. Specifically, we first describe the target of the fake news detection task. Then, the selected inside investigation information $\mathcal{II'} = \{ U_{positive}', U_{negative}' \}$ of current candidate news are followed, which augment the LLM's understanding of this task. Finally, we prompt the LLM to predict the result of current news and give its corresponding supportive explanation:
\begin{equation}
    \label{eq:inside_judge}
    P_i, R_i = \mathcal{F}\ (P_{ID} \ \uplus \ \mathcal{II'} \ \uplus \ S ),
\end{equation}
where $P_{ID}$ is the prompt instruction, $P_i$ refers to the prediction result, while $R_i$ denotes the corresponding explanation. You can move to Appendix~\ref{appendix:promot} for more details about this prompt.

\subsubsection{Outside Determination}
\label{sec:outside_determination}
Meanwhile, with the selected outside investigation information $\mathcal{OI'}$, we can derive the outside prediction, which is crucial for real-time news detection.

Similar to the design of Inside Judge, we describe the objective of fake news detection through an outside instruction, followed by the candidate news to be detected and the retrieved outside investigation documents $\mathcal{OI'}$. After that, we can derive the outside prediction $P_o$ and explanation $R_o$:
\begin{equation}
    \label{eq:onside_judge}
    P_o, R_o = \mathcal{F}\ (P_{OD} \ \uplus \ S \ \uplus \ \mathcal{OI'} ),
\end{equation}


\subsubsection{Integrated Determination}
\label{sec:integrated_determination}
With the predictions $P_i$, $P_o$ and their corresponding explanations $R_i$, $R_o$, the final outputs are obtained by jointly considering these two perspectives.

More specifically, if the two predictions are identical (i.e., $P_i = P_o$), we can directly derive the final prediction with high confidence. Nevertheless, if two results diverge, indicating a conflict between the Inside Determination and Outside Determination, we further propose a integrated selector to make a choice based on both sets of predictions and explanations:
\begin{equation}
\label{eq:pf}
    P_f =\mathcal{F}\ (P_{FD} \uplus S \uplus P_i \uplus R_i \uplus P_o \uplus R_o),
\end{equation}
where $P_f$ is the final inference result.


\section{Experiments}
\subsection{Experiment Setup}
\label{exp:setup}
\subsubsection{Datasets and Evaluation Metrics.}
We conduct experiments on two datasets, \textit{PolitiFact} and \textit{Gossipcop}, both of which are proposed in a benchmark called FakeNewsNet\cite{shu2020fakenewsnet}. PolitiFact consists of various political news, while Gossipcop is sourced from an entertainment story fact-checking website. For the few-shot setting, following the strategy employed in\cite{jiang2022fake,ma2023kapalm}, we randomly select $K \in (8, 32, 100)$ positive and negative news articles as the training set, respectively. More statistics about the datasets are illustrated in Table~\ref{tab:data}.

Given that the task focuses on detecting fake news, fake news articles are regarded as positive examples\cite{ma2023kapalm}. We further adopt the Accuracy (ACC) and F1-score~\cite{liu2023enhancing,liu2023techpat,ghanem2021fakeflow} as the evaluation metrics to measure classification performance.

\begin{table}[t]
    \centering
    \caption{Statistics of PolitiFact and Gossipcop datasets.}
    \renewcommand\arraystretch{1.15}
    \setlength{\tabcolsep}{5mm}{
    \begin{tabular}{c|c|c|c}
    \toprule
        \multicolumn{2}{c|}{Dataset} & \textbf{PolitiFact} & \textbf{Gossipcop} \\
        \midrule
        \multirow{3}{*}{Train}&
         \# True news & 8/32/100 & 8/32/100 \\
        & \# Fake news & 8/32/100 & 8/32/100 \\
        & \# Total news & 16/64/200 & 16/64/200 \\
        \midrule
        \multirow{3}{*}{Test}&
         \# True news & 120 & 3,200 \\
        & \# Fake news & 80 & 1,060 \\
        & \# Total news & 200 & 4,260 \\
    \bottomrule
    \end{tabular}}
    \label{tab:data}
    \vspace{-5mm}
\end{table}

\begin{table*}[t]
    \centering
    \caption{Experimental results of our proposed method on the PolitiFact and Gossipcop datasets.}
    \renewcommand\arraystretch{1.1}
    \setlength{\tabcolsep}{5mm}{
    \begin{tabular}{l|l|c|c|c|c|c|c }
    \toprule
    \multirow{2}{*}{Dataset} & \multirow{2}{*}{Methods}&
    \multicolumn{3}{c}{\textbf{ACC}} & \multicolumn{3}{|c}{\textbf{F-1 score}} \\
    \cmidrule{3-8}
     & & $K$=8 & $K$=32 & $K$=100 & $K$=8 & $K$=32 & $K$=100 \\
    \midrule
    \multirow{12}{*}{PolitiFact} 
    & \ding{172} PROPANEWS  & 40.00 & 43.50 & 40.00 & 57.14 & 58.30 & 57.14 \\
    & \ding{173} FakeFlow  & 61.00 & 62.50 & 63.50 & 44.29 & 47.55 & 48.95 \\
    & \ding{174} MDFEND & 65.50 & 64.00 & 71.50 & 62.30 & 64.36 & 69.84 \\
    & \ding{175} PSM  & 70.00 & 72.50 & 79.00 & 49.15 & 52.38 & 65.70  \\
    & \ding{176} KPL & 58.33 & 73.44 & 82.29 & 60.40 & 73.58 & 81.11 \\
    \cmidrule{2-8}
    & \ding{177} Auto-CoT & 49.50 & 58.00 & 64.00 & 53.88 & 58.00 & 55.00 \\
    & \ding{178} Zephyr  & 60.00 & 63.50 & 66.50 & 48.72 & 53.50 & 54.42  \\
    & \ding{179} ChatGLM-3  & 68.50 & 68.50 & 72.50 & 58.28 & 58.82 & 64.05  \\
    & \ding{180} LLama-3  & 69.50 & 70.50 & 69.00 & 63.91 & 65.09 & 64.00  \\
    & \circled{10} GPT-3.5  & 71.00 & 69.50 & 73.00 & 60.27 & 60.65 & 64.47  \\
    
    \cmidrule{2-8}
    & \circled{11} ARG  & 74.00 & 78.50 & 82.50 & 67.16 & 68.61 & 80.61  \\
    \cmidrule{2-8}
    & \textbf{DKFND (ours)}  & \textbf{87.00} & \textbf{88.00} & \textbf{89.00} & \textbf{82.43} & \textbf{83.78} & \textbf{85.33} \\

    \midrule

    \multirow{12}{*}{Gossipcop} 
    & \ding{172} PROPANEWS  & 24.88 & 25.40 & 24.88 & 39.85 & 39.97 & 39.85 \\
    & \ding{173} FakeFlow  & 57.89 & 58.26 & 57.28 & 26.60 & 27.66 & 28.18 \\
    & \ding{174} MDFEND & 41.27 & 56.08 & 63.73 & 40.20 & 42.06 & 44.52 \\
    & \ding{175} PSM  & 77.44 & 78.05 & 78.30 & 41.73 & 41.37 & 54.20  \\
    & \ding{176} KPL & 42.71 & 51.58 & 60.54 & 42.08 & 47.82 & 52.53 \\
    \cmidrule{2-8}
    & \ding{177} Auto-CoT & 52.44 & 45.54 & 48.73 & 28.46 & 34.72 & 33.46 \\
    & \ding{178} Zephyr  & 67.21 & 65.85 & 67.23 & 27.05 & 27.43 & 27.67  \\
    & \ding{179} ChatGLM-3  & 62.49 & 62.75 & 63.43 & 31.59 & 34.83 & 34.15  \\
    & \ding{180} LLama-3  & 65.96 & 65.85 & 66.17 & 30.89 & 35.07 & 31.74  \\
    & \circled{10} GPT-3.5  & 68.50 & 69.44 & 67.44 & 32.90 & 36.73 & 36.58  \\
    
    \cmidrule{2-8}
    & \circled{11} ARG  & 61.41 & 77.42 & 76.50 & 42.32 & 51.46 & 46.57  \\
    \cmidrule{2-8}
    & \textbf{DKFND (ours)}  & \textbf{82.37} & \textbf{82.18} & \textbf{82.56} & \textbf{55.22} & \textbf{55.17} & \textbf{56.78} \\
    
    \bottomrule
    \end{tabular}
    }
    \label{tab:main_result}
\end{table*}

\subsubsection{Implementation Details.}
\label{sec:inplementation}
In DKFND architecture, we utilize the \textit{zephyr-7b-beta}\cite{tunstall2023zephyr} model on Huggingface as the LLM. When running Zephyr, we adhere to the default parameter values provided by the official, where the sampling temperature is $0.70$, top\_k is $50$, and top\_p is $0.95$. The max\_new\_token is set to $256$, and do\_sample is set as $True$.

In the Detection Module (Section~\ref{section:detection}), the number of keywords to extract is set to $N = 5$. 

In the Inside Investigation part (Section~\ref{section:inside_investigation}), we employ the \textit{DeBERTa-base} model\cite{he2021deberta} from Transformers\cite{wolf2020transformers} as the representation model. The number of retrieved positive/negative nearest neighbors is set as $k = 5$.

In the Outside Investigation part (Section~\ref{section:outside_investigation}), we set the number of retrieved documents as $L=8$.

In the Inside Judge part (Section~\ref{section:inside_judge}), the number of positive/negative samples is set as $m = 2$.

In the Outside Judge part (Section~\ref{section:outside_judge}), we set the number of selected documents as $n=2$.

All experiments are conducted on a Linux server with two Tesla A100 GPUs.

\subsubsection{Benchmark Methods.}
For demonstrating DKFND's effectiveness, we compare it with the state-of-the-art few-shot fake news detection methods. According to the model architecture, they can be grouped into three categories, including traditional fake news detection methods (\ding{172} $\sim$ \ding{176}), LLM-based methods (\ding{177} $\sim$ \circled{10}), and hybrid methods (\circled{11}).

\begin{enumerate}
    \item[\ding{172}] \textbf{PROPANEWS}\cite{huang2023faking} proposes a framework for generating valuable training examples, beneficial to human-written situations.
    \item [\ding{173}] \textbf{FakeFlow}\cite{ghanem2021fakeflow} devises a model that detects fake news articles by integrating the ﬂow of affective information.
    \item [\ding{174}] \textbf{MDFEND}\cite{nan2021mdfend} incorporates the domain information through a gate mechanism to aggregate multiple representations.
    \item [\ding{175}] \textbf{PSM}\cite{ni2020improving} utilizes Propensity Score Matching to select decounfounded features, boosting detection performance.
    \item [\ding{176}] \textbf{KPL}\cite{jiang2022fake} incorporates external knowledge into the prompt representation process, thus achieving knowledgeable detection for fake news.
    \item [\ding{177}] \textbf{AutoCoT}\cite{zhang2022automatic} proposes an automatic chain-of-thought prompting method to construct demonstrations and reasoning chains.
    \item [\ding{178}] \textbf{Zephyr}\cite{tunstall2023zephyr} represents the advanced 7B model, which is optimized by the preference data from AI Feedback.
    \item [\ding{179}] \textbf{ChatGLM-3}\cite{zeng2022glm,du2022glm} is a series of pre-trained dialogue models, and we select the ChatGLM3-6B version. 
    \item [\ding{180}] \textbf{LLama-3}\cite{llama3modelcard} refers to the LLM proposed by Meta AI. We adopt its 8B version (Meta-Llama-3-8B-Instruct) for experiments.
    \item [\circled{10}] \textbf{GPT-3.5}\cite{ouyang2022training} is an advanced LLM developed by OpenAI. We leverage the API (version: gpt-3.5-turbo-0613) for in-context learning.
    \item [\circled{11}] \textbf{ARG}\cite{hu2024bad} designs an adaptive rationale guidance network, which integrates insights from both large language models and traditional methods.
\end{enumerate}

It is worth noting that, for these LLM-based baselines (\ding{178} $\sim$ \circled{10}), we adhere to the instruction prompt proposed by\cite{hu2024bad} to conduct in-context learning. Besides, due to the limitations of maximum tokens, we randomly select 6 samples as the demonstrations, which is more than the demonstration samples utilized in our DKFND model\footnote{As introduced in Section~\ref{sec:inplementation}, DKFND utilizes $n=2$ positive and negative samples as demonstrations, with a total number of 4.}, facilitating a fair comparison.

\subsection{Experimental Result}
The main results, presented in Table~\ref{tab:main_result}, indicate that our proposed DKFND model surpasses all baselines across various metrics, encompassing traditional, LLM-based and hybrid methods. This underscores the effectiveness of our design and the advantages of enhancing the LLM through both inside and outside perspectives. Furthermore, several notable phenomena emerge from these results:

Firstly, for most baselines and our DKFND model, the performance on the PolitiFact dataset exceeds that on the Gossipcop dataset, suggesting that Gossipcop presents greater difficulty. Specifically, PolitiFact consists of political news while Gossipcop pertains to the entertainment domain. This disparity is reasonable as political news typically exhibits more organized format and content, which facilitates the fake news detection process.
Secondly, with the increase of training instances ($K$), most traditional fake news detection methods (e.g., \ding{175} PSM) and hybrid methods (\circled{11} ARG) show improved performance. This is logical as more data enables better training of a supervised model, mitigating the lack of prior knowledge. However, an exception is observed in \ding{172} PROPANEWS, whose performance appears relatively unaffected by $K$. As introduced in Section~\ref{sec:inplementation}, different from other methods, PROPANEWS designs a data constructing strategy to supplement original training set. This significantly offsets the impact of training data quantity. Moreover, for most LLM-based methods (\ding{178} $\sim$ \circled{10}), as outlined in Section~\ref{sec:inplementation}, due to the limitation of maximum tokens, we all randomly select 6 samples as demonstrations. Hence, increasing the number of training instances does not substantially benefit the in-context learning of LLMs.
Thirdly, although our DKFND model also faces the constraint of maximum token limitations, the $k$NN retrieval mechanism and inside judge designs enhance the utilization of increased training data, thereby achieving a certain degree of improvements with higher $K$ values.
Fourthly, the hybrid method (\circled{11} ARG), benefiting from the joint modeling of traditional models and LLMs, obtains competitive performance. And compared to that, DKFND still maintains a significant advantage, particularly in scenarios with scarcer data.
These observations further demonstrate the effectiveness of our designs from multiple perspectives.

\subsection{Component Effectiveness Analysis}
In this subsection, we conduct ablation experiments to assess the effectiveness of different components within our model. Specifically, we first simplify the Integrated Determination part (Section \ref{sec:inside_determination}), removing the detailed explanation in promot $P_{FD}$ (Eq.(\ref{eq:pf})). On the basis of this, we design ablated variants from inside and outside perspectives.
\begin{itemize}
    \item \textbf{DKFND-Inside:} simplifying the design of Section \ref{section:inside_investigation} and \ref{section:inside_judge} by randomly selecting samples from training set as demonstrations.
    \item \textbf{DKFND-Outside:} simplifying the design of Section~\ref{section:outside_judge} by randomly selecting retrieved outside information.
    \item \textbf{DKFND-Both:} from both inside and outside perspectives, applying the above simplified processes to DKFND.
\end{itemize}

\begin{figure}[t]
    \centering
    \includegraphics[width=8.5cm]{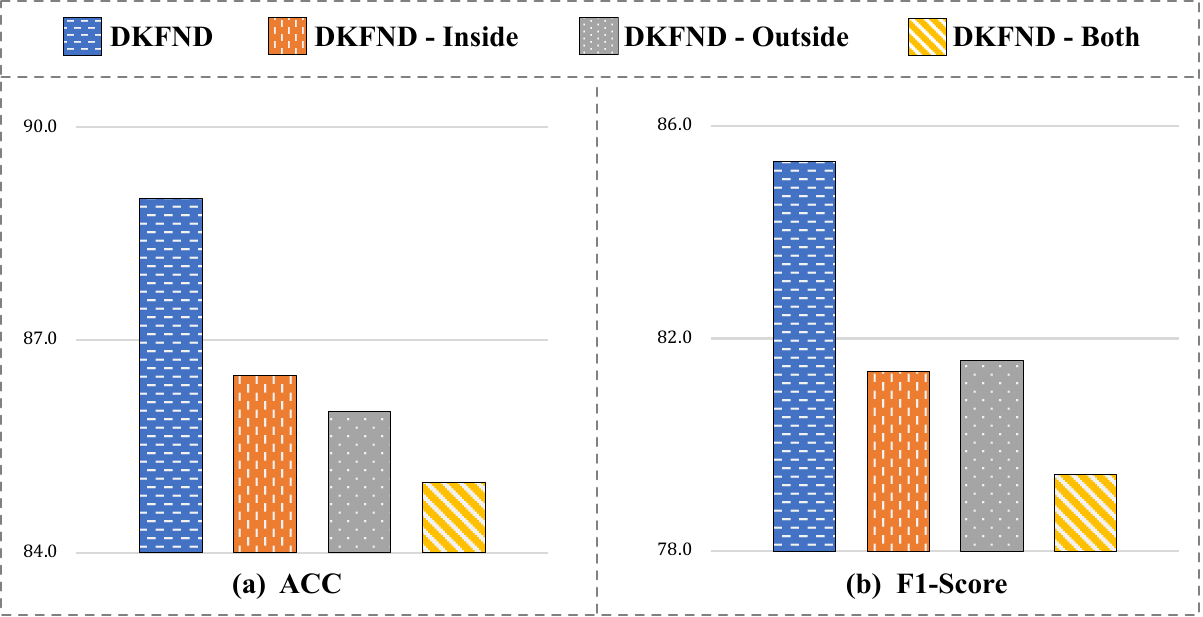}
    \caption{Ablation experiments on PolitiFact ($K$=100).}
    \vspace{-4mm}
    \label{fig:ablation}
\end{figure}

\begin{figure*}[t]
    \centering
    \includegraphics[width=0.9\textwidth]{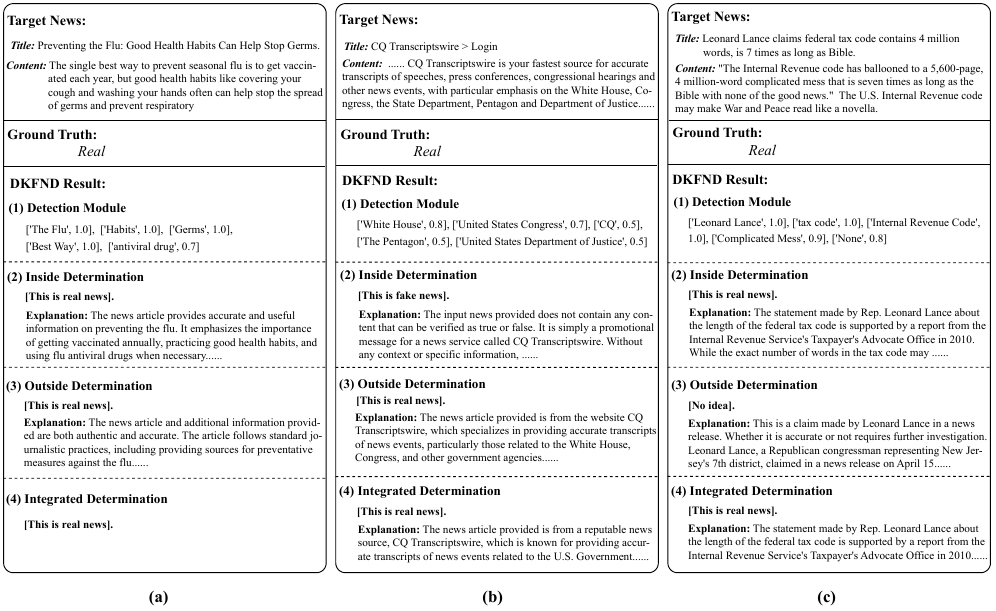}
    \vspace{-2mm}
    \caption{The case study of the DKFND model. They come from the PolitiFact dataset ($K$=100).}
    \vspace{-4mm}
    \label{fig:case}
\end{figure*}

The results on the PolitiFact ($K$=100) dataset are depicted in Figure~\ref{fig:ablation}. DKFND-Inside and DKFND-Outside both exhibit a significant performance drop compared to the full DKFND. Moreover, DAFND-Both, which applies simplifications from both inside and outside perspectives, demonstrates an even greater decline in detection performance. This observation highlights the effectiveness of our proposed dual-perspective design, particularly emphasizing the irreplaceable role of information selection and decision-making designs in DKFND.

Furthermore, considering that the backbone of DKFND is \ding{178} Zephyr\cite{tunstall2023zephyr}, we can jointly compare the results in Figure~\ref{fig:ablation} and the “\ding{178} Zephyr” line in Table~\ref{tab:main_result}. This comparison further reveals that all three ablation variants outperform Zephyr, demonstrating that even in its simplified forms, leveraging both internal and external perspectives for fake news detection yields significant performance gains. This finding further validates the motivation behind our dual-perspective design.

\subsection{Case Study}
To further illustrate the effectiveness of different modules in our model, we conduct a case study on the PolitiFact dataset. Specifically, Figure~\ref{fig:case} presents the input information (i.e., target news), the ground truth label, DKFND results (including the inside, outside and final results).

As shown in Figure~\ref{fig:case} (a), the Detection Module accurately identifies knowledge concepts relevant to the target news from KGs. Based on this, both Inside Determination and Outside Determination predict “\textit{[This is real news]}”. They are then fed into the Integrated Determination, which correctly predicts the label as \textit{[Real]}, consistent with the ground truth. In Figure~\ref{fig:case} (b), the Inside Determination incorrectly predicts \textit{[Fake]}, while the Outside Determination correctly predicts \textit{[Real]}. Conversely, in Figure Figure~\ref{fig:case} (c), the Inside Determination produces the correct prediction \textit{[Real]}, whereas the Outside Determination incorrectly classifies the news as \textit{[Fake]}. In both cases, thanks to the Integrated Determination, DKFND successfully reconciles conflicting predictions and ultimately arrives at the correct classification.

\section{Bad Case Analysis}
\label{sec:bad}
In this section, we illustrate the bad cases that DKFND struggles with, with a goal to analyze its shortcomings and possible improvement directions.

As illustrated in Figure~\ref{fig:badcase}, a particular failure occurs. From the explanation from Inside Determination and Outside Determination modules, they both classified it into \textit{Real} category because they haven't investigated clear negative information in contrast to the given news. More specifically, this bad case reveals two potential directions for improving our model: 
\begin{figure}[t]
    \centering
    \includegraphics[width=8cm]{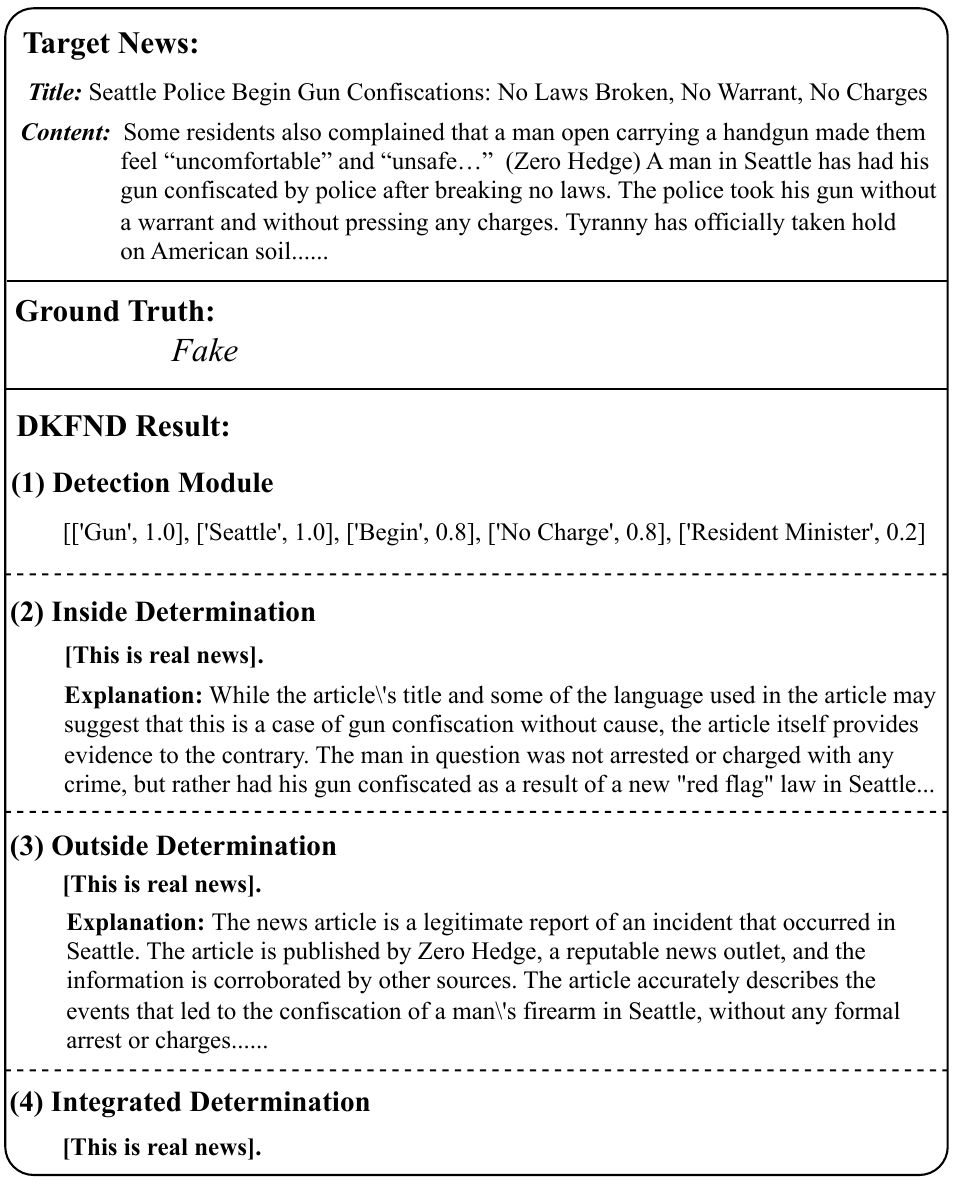}
    \caption{The bad case of DKFND on the Gossipcop dataset ($K$=100).}
    \vspace{-5mm}
    \label{fig:badcase}
\end{figure}
(1) For one thing, as introduced in Section~\ref{sec:introduction}, DKFND retrieves similar demonstrations from the training set in response to the Understanding Ambiguity problem. This relies on the assumption that valuable samples can be found in the training set, which is not always the case. As a consequence, the failure in Figure~\ref{fig:badcase} occurs. Therefore, how to discriminate and mitigate the impact of such circumstances is a crucial direction for improving the DKFND design. 
(2) For another, as introduced in Section~\ref{section:model}, we only employ the LLM to conduct inference, which cannot fully exploit LLMs' powerful capabilities. In some cases, although Outside Investigation retrieves valuable information, the prediction of Outside Judge still goes wrong. In fact, despite the great reasoning ability of these general LLMs, they are not competent in news-related domains and are not sufficiently familiar with the specific expression characteristics. Therefore, we would like to adopt the fine-tuning techniques to adapt LLMs for the news corpus in future, which we believe could bring a positive effect to our DKFND model.


\section{Limitation}
\label{appendix:limitation}
In our proposed DKFND method, we need to integrate the LLM (i.e., \textit{zephyr-7b-beta} introduced in Section~\ref{exp:setup}). Due to the large scale of LLMs, it tends to consume more computing resources and time compared to traditional baselines, such as PSM\cite{ni2020improving} and FakeFlow\cite{ghanem2021fakeflow}. Essentially speaking, LLMs contain a vast amount of knowledge, much of which may be unnecessary for fake news detection. Distilling useful knowledge so as to accelerate the inference remains a valuable and intriguing research direction.

Another limitation is that our current approach only employs LLMs for inference. Although we design precise prompts to implement in-context learning, it still cannot fully exploit the capabilities of LLMs due to the inherent gap between the natural language and the knowledge encoded in the model parameters. This also results in the bad case we analyzed in Section~\ref{sec:bad}.
In future work, we would like to explore the low resource scenario fine-tuning techniques (e.g., lora\cite{hu2021lora}) to adapt LLMs for few-shot fake news detection.
\section{Conclusions}
In this paper, we explored a motivated direction for few-shot fake news detection. We began by analyzing the limitations of current LLM-based detection methods, identifying two primary challenges: (1) Understanding Ambiguity and (2) Information Scarcity. To address these issues, we developed a Dual-perspective Knowledge-guided Fake News Detection (DKFND) model. In DKFND, a Detection Module was designed to identify knowledge concepts from given news. Then, we proposed an Investigation Module, a Judge Module to retrieve and select valuable information. Importantly, a Determination Module integrated predictions from both inside and outside perspectives to produce the final output. Finally, extensive experiments on two real-world datasets demonstrated the effectiveness of our method. We hope our work could lead to more future studies.


\section{Acknowledgements}
This research was partially supported by the National Natural Science Foundation of China (U23A20319, 62441239,2406303), Anhui Provincial Natural Science Foundation (No. 2308085QF229), Anhui Province Science and Technology Innovation Project (202423k09020010), the Fundamental Research Funds for the Central Universities (No. WK2150110034).

\bibliographystyle{ieeetr}
\bibliography{icdm}

\begin{thebibliography}{10}

\bibitem{shu2017fake}
K.~Shu, A.~Sliva, S.~Wang, J.~Tang, and H.~Liu, ``Fake news detection on social media: A data mining perspective,'' {\em ACM SIGKDD explorations newsletter}, vol.~19, no.~1, pp.~22--36, 2017.

\bibitem{hu2024bad}
B.~Hu, Q.~Sheng, J.~Cao, Y.~Shi, Y.~Li, D.~Wang, and P.~Qi, ``Bad actor, good advisor: Exploring the role of large language models in fake news detection,'' in {\em Proceedings of the AAAI Conference on Artificial Intelligence}, vol.~38, pp.~22105--22113, 2024.

\bibitem{gao2021making}
T.~Gao, A.~Fisch, and D.~Chen, ``Making pre-trained language models better few-shot learners,'' in {\em Proceedings of the 59th Annual Meeting of the Association for Computational Linguistics and the 11th International Joint Conference on Natural Language Processing (Volume 1: Long Papers)}, pp.~3816--3830, 2021.

\bibitem{ma2023kapalm}
J.~Ma, C.~Chen, C.~Hou, and X.~Yuan, ``Kapalm: Knowledge graph enhanced language models for fake news detection,'' in {\em Findings of the Association for Computational Linguistics: EMNLP 2023}, pp.~3999--4009, 2023.

\bibitem{horne2017just}
B.~Horne and S.~Adali, ``This just in: Fake news packs a lot in title, uses simpler, repetitive content in text body, more similar to satire than real news,'' in {\em Proceedings of the international AAAI conference on web and social media}, vol.~11, pp.~759--766, 2017.

\bibitem{jiang2022fake}
G.~Jiang, S.~Liu, Y.~Zhao, Y.~Sun, and M.~Zhang, ``Fake news detection via knowledgeable prompt learning,'' {\em Information Processing \& Management}, vol.~59, no.~5, p.~103029, 2022.

\bibitem{boissonneault2024fake}
D.~Boissonneault and E.~Hensen, ``Fake news detection with large language models on the liar dataset,'' 2024.

\bibitem{wang2024explainable}
B.~Wang, J.~Ma, H.~Lin, Z.~Yang, R.~Yang, Y.~Tian, and Y.~Chang, ``Explainable fake news detection with large language model via defense among competing wisdom,'' in {\em Proceedings of the ACM on Web Conference 2024}, pp.~2452--2463, 2024.

\bibitem{teo2024integrating}
T.~W. Teo, H.~N. Chua, M.~B. Jasser, and R.~T. Wong, ``Integrating large language models and machine learning for fake news detection,'' in {\em 2024 20th IEEE International Colloquium on Signal Processing \& Its Applications (CSPA)}, pp.~102--107, IEEE, 2024.

\bibitem{ghanem2021fakeflow}
B.~Ghanem, S.~P. Ponzetto, P.~Rosso, and F.~Rangel, ``Fakeflow: Fake news detection by modeling the flow of affective information,'' in {\em Proceedings of the 16th Conference of the European Chapter of the Association for Computational Linguistics: Main Volume}, pp.~679--689, 2021.

\bibitem{dun2021kan}
Y.~Dun, K.~Tu, C.~Chen, C.~Hou, and X.~Yuan, ``Kan: Knowledge-aware attention network for fake news detection,'' in {\em Proceedings of the AAAI conference on artificial intelligence}, vol.~35, pp.~81--89, 2021.

\bibitem{hu2021compare}
L.~Hu, T.~Yang, L.~Zhang, W.~Zhong, D.~Tang, C.~Shi, N.~Duan, and M.~Zhou, ``Compare to the knowledge: Graph neural fake news detection with external knowledge,'' in {\em Proceedings of the 59th Annual Meeting of the Association for Computational Linguistics and the 11th International Joint Conference on Natural Language Processing (Volume 1: Long Papers)}, pp.~754--763, 2021.

\bibitem{huang2023faking}
K.-H. Huang, K.~Mckeown, P.~Nakov, Y.~Choi, and H.~Ji, ``Faking fake news for real fake news detection: Propaganda-loaded training data generation,'' in {\em Proceedings of the 61st Annual Meeting of the Association for Computational Linguistics (Volume 1: Long Papers)}, pp.~14571--14589, 2023.

\bibitem{hoffmann2022training}
J.~Hoffmann, S.~Borgeaud, A.~Mensch, E.~Buchatskaya, T.~Cai, E.~Rutherford, D.~d.~L. Casas, L.~A. Hendricks, J.~Welbl, A.~Clark, {\em et~al.}, ``Training compute-optimal large language models,'' {\em arXiv preprint arXiv:2203.15556}, 2022.

\bibitem{openai2023gpt4}
OpenAI, ``Gpt-4 technical report,'' 2023.

\bibitem{llama3modelcard}
AI@Meta, ``Llama 3 model card,'' 2024.

\bibitem{tunstall2023zephyr}
L.~Tunstall, E.~Beeching, N.~Lambert, N.~Rajani, K.~Rasul, Y.~Belkada, S.~Huang, L.~von Werra, C.~Fourrier, N.~Habib, {\em et~al.}, ``Zephyr: Direct distillation of lm alignment,'' {\em arXiv preprint arXiv:2310.16944}, 2023.

\bibitem{brown2020language}
T.~Brown, B.~Mann, N.~Ryder, M.~Subbiah, J.~D. Kaplan, P.~Dhariwal, A.~Neelakantan, P.~Shyam, G.~Sastry, A.~Askell, {\em et~al.}, ``Language models are few-shot learners,'' {\em Advances in neural information processing systems}, vol.~33, pp.~1877--1901, 2020.

\bibitem{lewis2020retrieval}
P.~Lewis, E.~Perez, A.~Piktus, F.~Petroni, V.~Karpukhin, N.~Goyal, H.~K{\"u}ttler, M.~Lewis, W.-t. Yih, T.~Rockt{\"a}schel, {\em et~al.}, ``Retrieval-augmented generation for knowledge-intensive nlp tasks,'' {\em Advances in Neural Information Processing Systems}, vol.~33, pp.~9459--9474, 2020.

\bibitem{izacard2021leveraging}
G.~Izacard and {\'E}.~Grave, ``Leveraging passage retrieval with generative models for open domain question answering,'' in {\em Proceedings of the 16th Conference of the European Chapter of the Association for Computational Linguistics: Main Volume}, pp.~874--880, 2021.

\bibitem{borgeaud2022improving}
S.~Borgeaud, A.~Mensch, J.~Hoffmann, T.~Cai, E.~Rutherford, K.~Millican, G.~B. Van Den~Driessche, J.-B. Lespiau, B.~Damoc, A.~Clark, {\em et~al.}, ``Improving language models by retrieving from trillions of tokens,'' in {\em International conference on machine learning}, pp.~2206--2240, PMLR, 2022.

\bibitem{izacard2023atlas}
G.~Izacard, P.~Lewis, M.~Lomeli, L.~Hosseini, F.~Petroni, T.~Schick, J.~Dwivedi-Yu, A.~Joulin, S.~Riedel, and E.~Grave, ``Atlas: Few-shot learning with retrieval augmented language models,'' {\em Journal of Machine Learning Research}, vol.~24, no.~251, pp.~1--43, 2023.

\bibitem{honnibal2017spacy}
M.~Honnibal and I.~Montani, ``spacy 2: Natural language understanding with bloom embeddings, convolutional neural networks and incremental parsing,'' {\em To appear}, vol.~7, no.~1, pp.~411--420, 2017.

\bibitem{liu2022makes}
J.~Liu, D.~Shen, Y.~Zhang, W.~B. Dolan, L.~Carin, and W.~Chen, ``What makes good in-context examples for gpt-3?,'' in {\em Proceedings of Deep Learning Inside Out (DeeLIO 2022)}, pp.~100--114, 2022.

\bibitem{rubin2022learning}
O.~Rubin, J.~Herzig, and J.~Berant, ``Learning to retrieve prompts for in-context learning,'' in {\em Proceedings of the 2022 Conference of the North American Chapter of the Association for Computational Linguistics: Human Language Technologies}, pp.~2655--2671, 2022.

\bibitem{khandelwal2019generalization}
U.~Khandelwal, O.~Levy, D.~Jurafsky, L.~Zettlemoyer, and M.~Lewis, ``Generalization through memorization: Nearest neighbor language models,'' in {\em International Conference on Learning Representations}, 2019.

\bibitem{yoran2023answering}
O.~Yoran, T.~Wolfson, B.~Bogin, U.~Katz, D.~Deutch, and J.~Berant, ``Answering questions by meta-reasoning over multiple chains of thought,'' in {\em Proceedings of the 2023 Conference on Empirical Methods in Natural Language Processing}, pp.~5942--5966, 2023.

\bibitem{paranjape2023art}
B.~Paranjape, S.~Lundberg, S.~Singh, H.~Hajishirzi, L.~Zettlemoyer, and M.~T. Ribeiro, ``Art: Automatic multi-step reasoning and tool-use for large language models,'' {\em arXiv preprint arXiv:2303.09014}, 2023.

\bibitem{trouillon2016complex}
T.~Trouillon, J.~Welbl, S.~Riedel, {\'E}.~Gaussier, and G.~Bouchard, ``Complex embeddings for simple link prediction,'' in {\em International conference on machine learning}, pp.~2071--2080, PMLR, 2016.

\bibitem{broscheit2020libkge}
S.~Broscheit, D.~Ruffinelli, A.~Kochsiek, P.~Betz, and R.~Gemulla, ``Libkge-a knowledge graph embedding library for reproducible research,'' in {\em Proceedings of the 2020 conference on empirical methods in natural language processing: system demonstrations}, pp.~165--174, 2020.

\bibitem{patil2021comparative}
A.~Patil, J.~Pamnani, and D.~Pawade, ``Comparative study of google search engine optimization algorithms: Panda, penguin and hummingbird,'' in {\em 2021 6th International Conference for Convergence in Technology (I2CT)}, pp.~1--5, IEEE, 2021.

\bibitem{shu2020fakenewsnet}
K.~Shu, D.~Mahudeswaran, S.~Wang, D.~Lee, and H.~Liu, ``Fakenewsnet: A data repository with news content, social context and spatiotemporal information for studying fake news on social media,'' {\em Journal on big data}, vol.~8, no.~3, 2020.

\bibitem{liu2023enhancing}
Y.~Liu, K.~Zhang, Z.~Huang, K.~Wang, Y.~Zhang, Q.~Liu, and E.~Chen, ``Enhancing hierarchical text classification through knowledge graph integration,'' in {\em Findings of the association for computational linguistics: ACL 2023}, pp.~5797--5810, 2023.

\bibitem{liu2023techpat}
Y.~Liu, H.~Wu, Z.~Huang, H.~Wang, Y.~Ning, J.~Ma, Q.~Liu, and E.~Chen, ``Techpat: technical phrase extraction for patent mining,'' {\em ACM Transactions on Knowledge Discovery from Data}, vol.~17, no.~9, pp.~1--31, 2023.

\bibitem{he2021deberta}
P.~He, X.~Liu, J.~Gao, and W.~Chen, ``Deberta: Decoding-enhanced bert with disentangled attention,'' in {\em International Conference on Learning Representations}, 2021.

\bibitem{wolf2020transformers}
T.~Wolf, L.~Debut, V.~Sanh, J.~Chaumond, C.~Delangue, A.~Moi, P.~Cistac, T.~Rault, R.~Louf, M.~Funtowicz, {\em et~al.}, ``Transformers: State-of-the-art natural language processing,'' in {\em Proceedings of the 2020 conference on empirical methods in natural language processing: system demonstrations}, pp.~38--45, 2020.

\bibitem{nan2021mdfend}
Q.~Nan, J.~Cao, Y.~Zhu, Y.~Wang, and J.~Li, ``Mdfend: Multi-domain fake news detection,'' in {\em Proceedings of the 30th ACM International Conference on Information \& Knowledge Management}, pp.~3343--3347, 2021.

\bibitem{ni2020improving}
B.~Ni, Z.~Guo, J.~Li, and M.~Jiang, ``Improving generalizability of fake news detection methods using propensity score matching,'' {\em arXiv preprint arXiv:2002.00838}, 2020.

\bibitem{zhang2022automatic}
Z.~Zhang, A.~Zhang, M.~Li, and A.~Smola, ``Automatic chain of thought prompting in large language models,'' {\em arXiv preprint arXiv:2210.03493}, 2022.

\bibitem{zeng2022glm}
A.~Zeng, X.~Liu, Z.~Du, Z.~Wang, H.~Lai, M.~Ding, Z.~Yang, Y.~Xu, W.~Zheng, X.~Xia, {\em et~al.}, ``Glm-130b: An open bilingual pre-trained model,'' {\em arXiv preprint arXiv:2210.02414}, 2022.

\bibitem{du2022glm}
Z.~Du, Y.~Qian, X.~Liu, M.~Ding, J.~Qiu, Z.~Yang, and J.~Tang, ``Glm: General language model pretraining with autoregressive blank infilling,'' in {\em Proceedings of the 60th Annual Meeting of the Association for Computational Linguistics (Volume 1: Long Papers)}, pp.~320--335, 2022.

\bibitem{ouyang2022training}
L.~Ouyang, J.~Wu, X.~Jiang, D.~Almeida, C.~Wainwright, P.~Mishkin, C.~Zhang, S.~Agarwal, K.~Slama, A.~Ray, {\em et~al.}, ``Training language models to follow instructions with human feedback,'' {\em Advances in Neural Information Processing Systems}, vol.~35, pp.~27730--27744, 2022.

\bibitem{hu2021lora}
E.~J. Hu, Y.~Shen, P.~Wallis, Z.~Allen-Zhu, Y.~Li, S.~Wang, L.~Wang, and W.~Chen, ``Lora: Low-rank adaptation of large language models,'' {\em arXiv preprint arXiv:2106.09685}, 2021.

\end{thebibliography}

\appendix

\subsection{Outside Investigatation Illustration}
\label{appendix:outside_search}
\begin{figure}[ht]
    \centering
    \includegraphics[width=7cm]{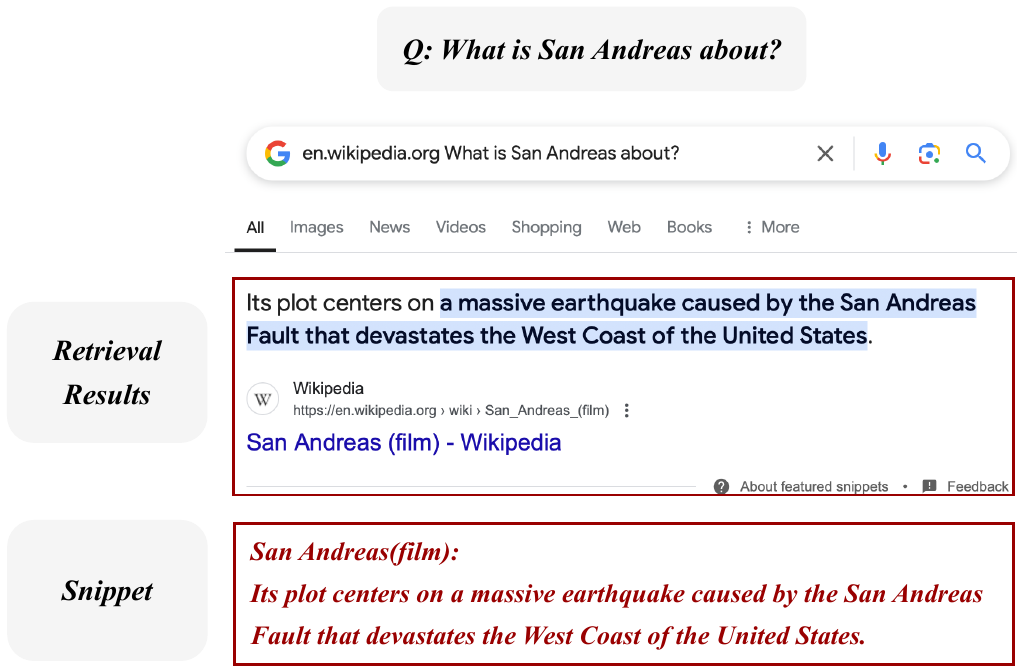}
    \caption{Illustration of the Google search process.}
    \label{fig:outside_search}
    \vspace{-4mm}
\end{figure}

For better illustrate the search process of the outside investigation, we give an example in Figure~\ref{fig:outside_search}. You can also refer to our released codes for more technical details.
\subsection{Prompts of DKFND}
\label{appendix:promot}
\begin{figure*}[t]
    \centering
    \includegraphics[width=\textwidth]{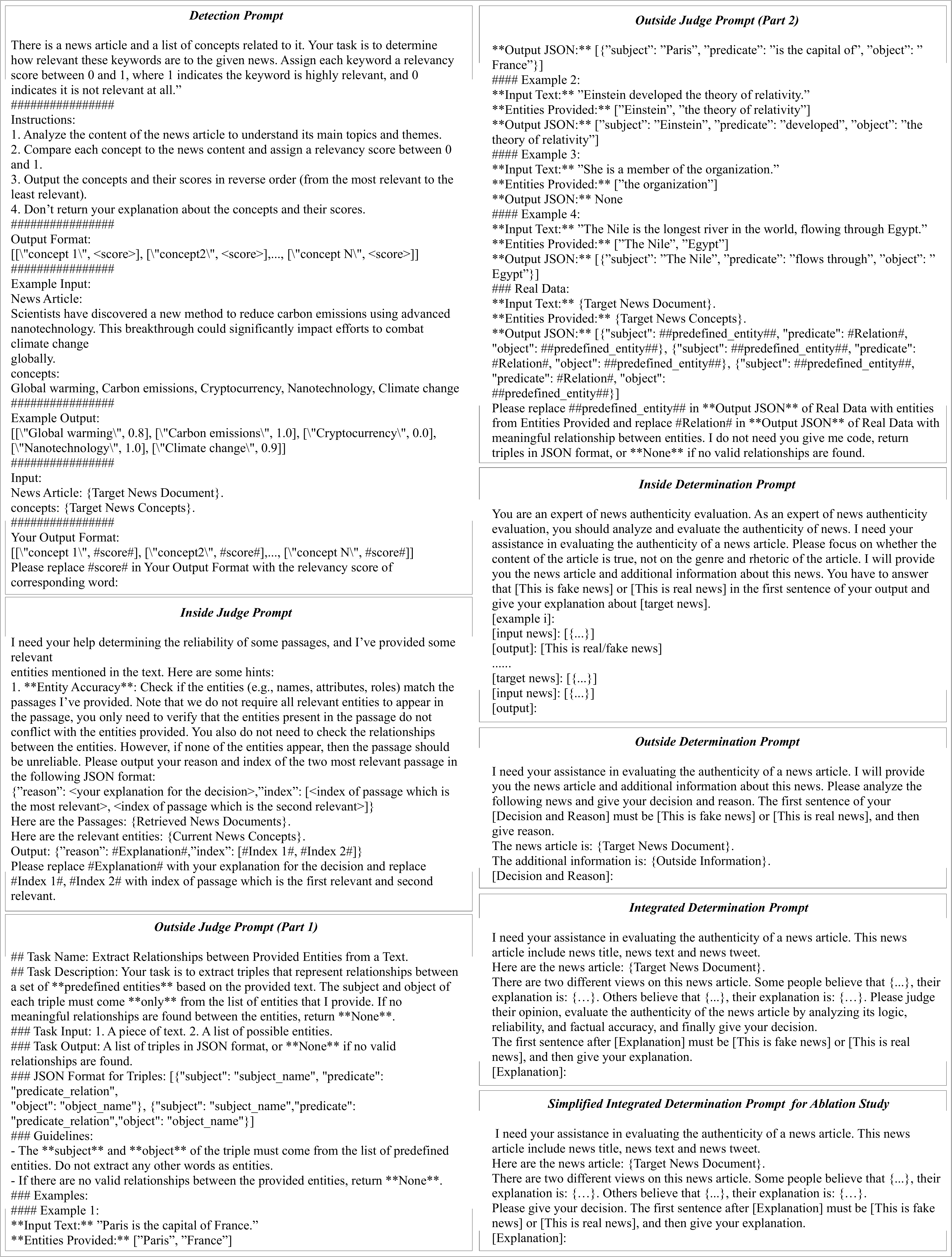}
    \caption{Prompts in our DKFND methodology.}
    \label{fig:prompts}
\end{figure*}

\end{document}